\documentclass[conference]{IEEEtran}
\IEEEoverridecommandlockouts
\usepackage{cite}
\usepackage{amsmath,amssymb,amsfonts}
\usepackage{algorithmic}
\usepackage{graphicx}
\usepackage{textcomp}
\usepackage{xcolor}
\usepackage{amsmath}
\usepackage{amsfonts}
\usepackage{amssymb}
\usepackage{xspace}
\usepackage{multicol}
\usepackage{multirow}
\usepackage{enumerate}
\usepackage{subcaption}
\usepackage{bbding}
\usepackage{enumitem}

\newcommand{\pbppo}{{\rm $\text{P}^3\text{O}$}\xspace}
\newcommand{\promptTR}{{{prompt-transformer}}\xspace}
\newcommand{\promptTRUp}{{{Prompt-Transformer}}\xspace}
\newcommand{\conversion}{{{representation}}\xspace}
\newcommand{\conversionUp}{{{Representation}}\xspace}
\newcommand{\parsection}[1]{\vspace{0.5mm}\noindent\textbf{#1.}~}

\def\BibTeX{{\rm B\kern-.05em{\sc i\kern-.025em b}\kern-.08em
    T\kern-.1667em\lower.7ex\hbox{E}\kern-.125emX}}
\begin{document}

\title{$\text{P}^3\text{O}$: Transferring Visual Representations for Reinforcement Learning via Prompting\\
\thanks{This work was supported by Guangdong Province R\&D Program 2020B0909050001, Anhui Province Development and Reform Commission 2020 New Energy Vehicle Industry Innovation Development Project and 2021 New Energy and Intelligent Connected Vehicle Innovation Project, and Shenzhen Yijiahe Technology R\&D Co., Ltd. * The corresponding author.}
}

\makeatletter
\newcommand{\linebreakand}{%
  \end{@IEEEauthorhalign}
  \hfill\mbox{}\par
  \mbox{}\hfill\begin{@IEEEauthorhalign}
}
\makeatother

\author{\IEEEauthorblockN{Guoliang You}
\IEEEauthorblockA{\textit{School of Computer Science} \\ \textit{and Technology} \\
\textit{University of Science and Technology} \\ \textit{of China}\\
Anhui, China \\
glyou@mail.ustc.edu.cn}
\and
\IEEEauthorblockN{Xiaomeng Chu}
\IEEEauthorblockA{\textit{School of Computer Science} \\ \textit{and Technology} \\
\textit{University of Science and Technology} \\ \textit{of China}\\
Anhui, China \\
cxmeng@mail.ustc.edu.cn}
\and
\IEEEauthorblockN{Yifan Duan}
\IEEEauthorblockA{\textit{School of Computer Science} \\ \textit{and Technology} \\
\textit{University of Science and Technology} \\ \textit{of China}\\
Anhui, China \\
dyf0202@mail.ustc.edu.cn}
\and 
\IEEEauthorblockN{Jie Peng}
\IEEEauthorblockA{\textit{School of Computer Science} \\ \textit{and Technology} \\
\textit{University of Science and Technology} \\ \textit{of China}\\
Anhui, China \\
pengjieb@mail.ustc.edu.cn}
\and
\IEEEauthorblockN{Jianmin Ji}
\IEEEauthorblockA{\textit{School of Computer Science} \\ \textit{and Technology} \\
\textit{University of Science and Technology} \\ \textit{of China}\\
Anhui, China \\
jianmin@ustc.edu.cn}
\and
\IEEEauthorblockN{Yu Zhang}
\IEEEauthorblockA{\textit{School of Computer Science} \\ \textit{and Technology} \\
\textit{University of Science and Technology} \\ \textit{of China}\\
Anhui, China \\
yuzhang@ustc.edu.cn}
\linebreakand
\IEEEauthorblockN{Yanyong Zhang*}
\IEEEauthorblockA{\textit{School of Computer Science} \\ \textit{and Technology} \\
\textit{University of Science and Technology} \\ \textit{of China}\\
Anhui, China \\
yanyongz@ustc.edu.cn}
}

\maketitle

\begin{abstract}
It is important for deep reinforcement learning (DRL) algorithms to transfer their learned policies to new environments that have different visual inputs.
In this paper, we introduce \textbf{P}rompt based \textbf{P}roximal \textbf{P}olicy \textbf{O}ptimization (\textbf{\pbppo}), a three-stage DRL algorithm that transfers visual representations from a target to a source environment by applying prompting. The process of \pbppo consists of three stages: pre-training, prompting, and predicting.
In particular, we specify a \promptTR for \conversion conversion and propose a two-step training process to train the \promptTR for the target environment, while the rest of the DRL pipeline remains unchanged. 
We implement \pbppo and evaluate it on the OpenAI CarRacing video game.
The experimental results show that \pbppo outperforms the  state-of-the-art visual transferring schemes. In particular, \pbppo allows the learned policies to perform well in environments with different visual inputs, which is much more effective than retraining the policies in these environments.
\end{abstract}

\begin{IEEEkeywords}
Visual Transfer, Reinforcement Learning, Imitation Learning, Prompting Method
\end{IEEEkeywords}
\vspace{-3pt}

\section{Introduction}

\begin{figure*}[t]
	\centering
    \includegraphics[width = 0.98\linewidth]{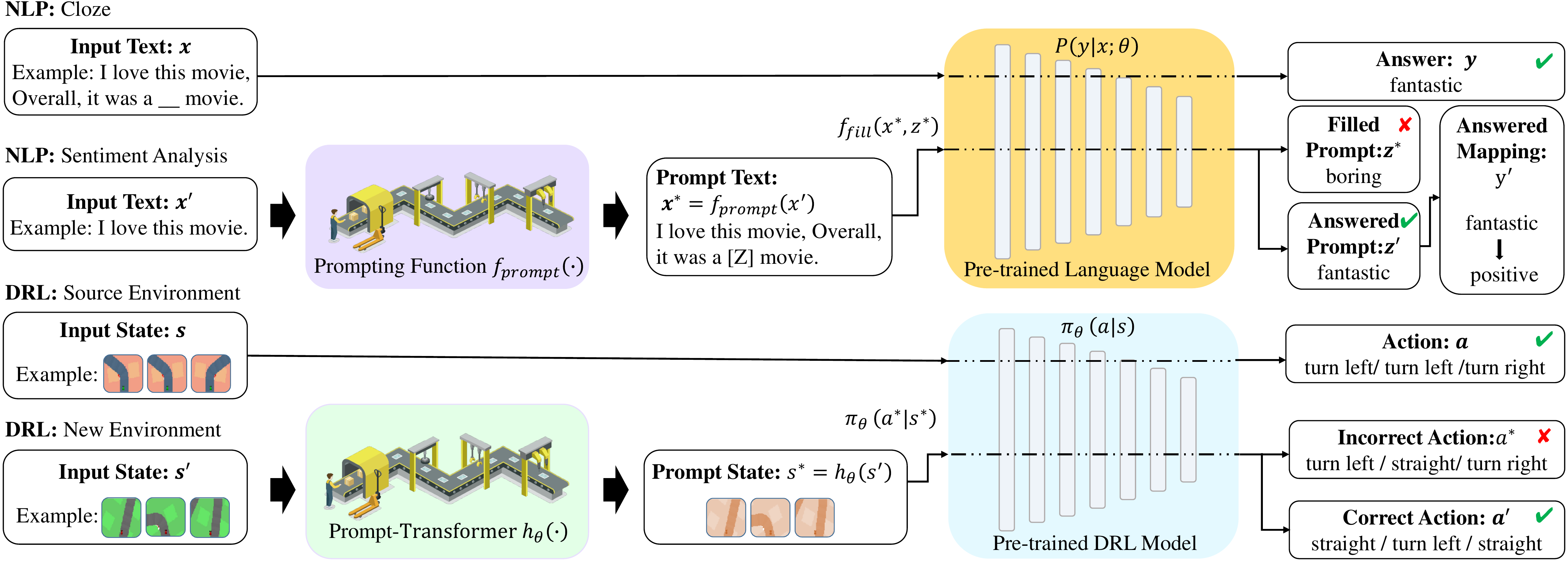}
	\caption{The correspondence between \pbppo (lower) and the prompting method in NLP (upper). The prompt-transformer, prompt state, incorrect action, and correct action in our pipeline are mapped to prompt-function, prompt text, filled prompt, and answered prompt in NLP, respectively.}
	\label{fig:PromptNlpDrl}
\end{figure*} 

Deep Reinforcement Learning (DRL) has been applied to a large set of applications, including games, robotics, self-driving~\cite{schulman2017proximal,mnih2013playing,drlRobot,DRLSelfCar}, etc. 
However, it is challenging for DRL algorithms to transfer these pre-trained models to new environments with different visual inputs~\cite {tobin2017domain}.
For example, the performance of the pre-trained model is often reduced or even completely collapsed in a new environment, even if there are only minor differences from the source environments, such as adding irregular shapes or changing colors~\cite{ShaniGamrian2019TransferLF}.
Retraining or fine-tuning the model from scratch for a new environment is usually expensive~\cite{fickinger2021scalable}.
In many cases, the reward function and network structure designed for the source environment are unsuitable for the new one, and hence the policies obtained by retraining or fine-tuning also perform poorly.

To address this challenge, some studies focus on extracting cross-domain features for training~\cite{DBLP:conf/aaai/XingNCZNK21,DBLP:conf/iclr/0001MCGL21}, either general features or task-specific features. Some focus on transforming between the source and target domains~\cite{ShaniGamrian2019TransferLF,you2017virtual, Roy_Konidaris_2021}. These methods require complex network structures and loss functions, as well as large amounts of data; they sometimes require re-completing the entire DRL training process,  which is costly and time-consuming.
Indeed, these methods do not use the pre-trained model knowledge from the source domain, but rather retrain or fine-tune the model.

In natural language processing (NLP), the prompting method has been incorporated for various tasks and led to promising results~\cite{liu2021pre}. 
In general, the prompt function typically adds prompt tokens for adapting pre-trained language models to achieve satisfactory performance in downstream tasks.
In this paper, we propose Prompt Based Proximal Policy Optimization (\pbppo), a three-stage DRL algorithm that uses prompt to transfer visual representations from target to source tasks, leveraging pre-trained models to achieve similar performance in the target environment. The three stages in \pbppo can be summarized as (1) pre-training, (2) prompting, and (3) predicting, where the correspondence between ``prompting" in DRL and that in NLP is shown in Fig.~\ref{fig:PromptNlpDrl}.

Specifically, in the prompting process, we use a multi-layer convolutional neural network to build \promptTR to fit the prompting function. 
We expect to use DRL's continuous optimization to learn \promptTR's parameters, but many invalid explorations cause slow or failed convergence. Imitation learning with expert knowledge can reduce ineffective exploration and speed up learning.
Therefore, we divide the training of \promptTR into two steps. In step~1, we use mini data in the target environment to initialize \promptTR with imitation learning, so that it guides prompt-transformer to form reasonable initialization weights and reduce invalid explorations in step~2. Then, in step~2, DRL continuously optimizes \promptTR by collecting observation data from target environments and trying to obtain output actions for higher rewards. It is worth noting that in both steps, we only update the weights of the \promptTR and freeze the other weights pre-trained in the source environment.
We conduct the experiments on the CarRacing video game in OpenAI Gym.
The results show that \pbppo can apply pre-trained model knowledge to a target environment, achieving state-of-the-art performance in visual representation transfer.

In summary, our main contributions are as follows:
\begin{itemize}
  \item We use the prompting method to solve the visual representation transfer problem in DRL, allowing the model pre-trained in the source environment to fully recover its performance in the target environment.
  \item We propose \pbppo, a three-stage DRL algorithm that specifies a \promptTR for the representation transfer in a prompting method. It enables source knowledge to be applied to target environments with different visual representations.
  \item Experiments on the CarRacing video game show that \pbppo outperforms the state-of-the-art method in most environments, with improved training efficiency and algorithm confidence interval.
\end{itemize}

\section{Related Work}
\begin{figure*}[t]
	\centering
    \includegraphics[width = 0.98\linewidth]{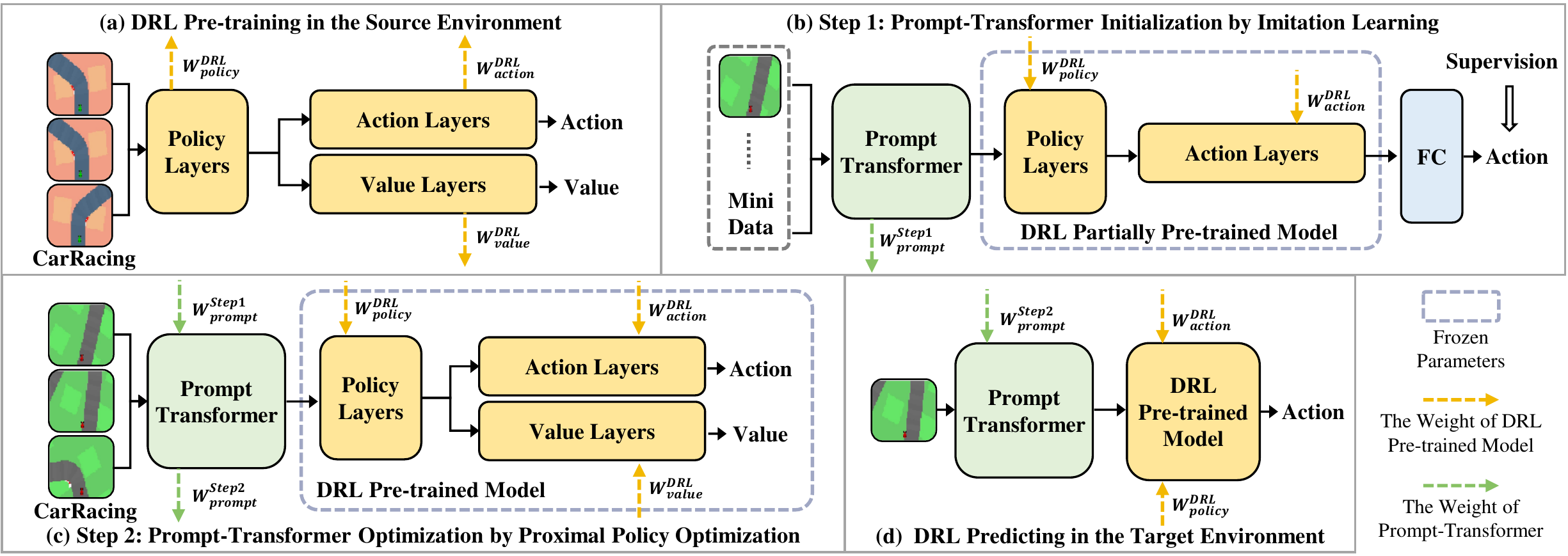}
	\caption{\label{mainArc}The overview of \pbppo. The DRL network is composed of policy, action, and value layers. 
	(a) We train the model in the source environment as the pre-trained model.
	(b) The imitation learning network in step~1, which is used to initialize the \promptTR using mini expert data from target environments. 
	(c) The weights of the prompt-transformer initialized in step~1 are optimized by DRL in target environments in step~2.
	(d) The optimal \promptTR optimized in step~2 is combined with the pre-trained model in (a) to obtain the highest-scoring action output in the target environment.}
\end{figure*} 

\subsection{Deep Reinforcement Learning}
A model trained in DRL may perform poorly in a new environment. Yurong et al.~\cite{you2017virtual} use Generative Adversarial Networks (GANs) to transform the simulated domain into the real domain. Gamrian et al.~\cite{ShaniGamrian2019TransferLF} use Unaligned GANs to transform the target environment representation back to the source environment without paired data. Xing et al.~\cite{DBLP:conf/aaai/XingNCZNK21} use Cycle-Consistent VAE to extract domain-general features in different environments for transfer. Roy et al.~\cite{Roy_Konidaris_2021} minimize the Wasserstein-1 distance of the features between the source and the target to learn a common latent space.

\subsection{Imitation Learning}
Imitation learning~\cite{ho2016generative} is used in fields such as joint motion~\cite{NathanRatliff2007ImitationLF}, robot manipulation~\cite{icra/Johns21}, and autonomous driving~\cite{MariuszBojarski2016EndTE}, \emph{etc}. It learns quickly through expert demonstrations but cannot outperform human experts. Collecting expert data is complex, and the model's performance degrades in new environments.

\subsection{Prompting Method}
The development of NLP can be grouped into four paradigms~\cite{liu2021pre}: Fully Supervised Learning (Non-Neural Network), Fully Supervised Learning (Neural Network), ``Pre-train, Fine-tune", and ``Pre-train, Prompt, Prediction". ``Pre-train, Fine-tune" has gained popularity and has produced many pre-trained language models (PLM)~\cite{devlin2018bert,brown2020language,raffel2019exploring}, which can be applied to new tasks after fine-tuning, but the increasing size of PLMs requires more hardware and data. The ``Pre-train, Prompt, Prediction" paradigm allows tasks to adapt to the PLM without fine-tuning, speeding up learning and reducing training difficulty and parameters.

\section{The \pbppo Design}
In this section, we explain the detailed design of \pbppo. Fig.~\ref{mainArc} shows the pipeline of our method, which consists of the following three stages: 
\begin{enumerate}[itemindent=1pt]
\item[(1)] DRL pre-training. This stage is to use the PPO algorithm to obtain a high-performance pre-trained model in the source environment.
\item[(2)] Prompting. This stage is to optimize the \promptTR for representation transfer, which is composed of the following two steps:
\begin{enumerate}[itemindent=4pt]
\item[(a)] Step~1: \promptTR initialization by imitation learning. The first step is to use the mini data in the target environments to initialize \promptTR with imitation learning.
\item[(b)] Step~2: \promptTR optimization by proximal policy optimization. The second step is to optimize \promptTR in DRL by collecting data from target environments to restore the original performance in the source environment.
\end{enumerate}
\item[(3)] DRL predicting. This stage is to use the optimal \promptTR to convert the \conversion from target to source and input it into the pre-trained model, which then predicts the correct actions.
\end{enumerate}

\subsection{DRL Pre-training in the Source Environment} 
\label{pre-trainedDrl}
We specify the CarRacing video game as a Markov Decision Process (MDP) problem.
The agent is pre-trained in the source environment using the PPO algorithm~\cite{schulman2017proximal}, as shown in Fig.~\ref{mainArc}a). 
Specifically, PPO parameterizes a policy $\pi_{\theta}(a \mid s)$ and a value function $V_{\theta}(s)$, where $a$ is action and $s$ are the observation data in the source environments. The training objective of the policy network in PPO is to minimize the policy loss:
\begin{equation}
\begin{split}
\label{eq1}
    \mathcal{L}_{\text {policy }}^{DRL}=&-\hat{\mathbb{E}}\left[\min \left(r_{\theta} \hat{A}{(s,a)},\right.\right.\\ 
    &\left.\left.\operatorname{clip}\left(r_{\theta}, 1-\epsilon, 1+\epsilon\right) \hat{A}{(s,a)}\right]\right.
\end{split}
\end{equation}
where $\hat{\mathbb{E}}$ is the empirical expected value and $\hat{A}$ is the expected advantage. $r(\theta)$ is the ratio of the current policy to the previous policy, and $\epsilon$ is the hyper-parameter. The clip function is defined as $\operatorname{clip}(\mu, \alpha, \beta) = \max(\min(\mu, \beta), \alpha)$.
The training objective of the PPO value network is to minimize the mean squared error, where $R$ is the return value:
\begin{equation}
\begin{aligned}
\label{eq2}
    \mathcal{L}_{\text {value }}^{DRL}=\frac{1}{n} \sum_{i=1}^{n}(V(s)-R)^{2}
\end{aligned}
\end{equation}
Finally, the policy trained in the source environments serves as the pre-trained model $\pi_{\theta}(a \mid s)$, with weights ${W_{policy}^{DRL}}$, ${W_{action}^{DRL}}$, and ${W_{value}^{DRL}}$ reused without retraining or fine-tuning in step~1 and step~2.

\subsection{\promptTRUp for \conversionUp Conversion in Target Environments}

\begin{figure}[t]
    \centering
    \includegraphics[width = 0.98\linewidth]{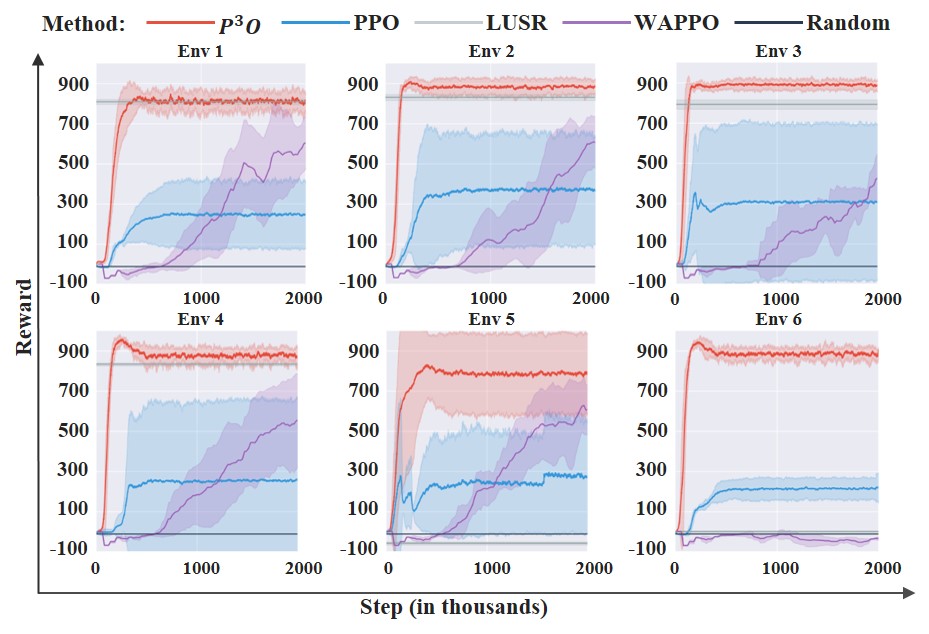}
    \caption{\label{fig:PromptBasedDRLexp1}The comparison of the reward curves for the training by \pbppo (in \textcolor[RGB]{224,51,46}{red}), PPO (in \textcolor[RGB]{40,129,210}{blue}), LUSR (in \textcolor[RGB]{140,155,159}{gray}), WAPPO (in \textcolor[RGB]{138,67,168}{purple}), and random policy (in \textcolor[RGB]{58,74,96}{black}) in the same new environments from $Env1\sim Env6$.}\label{PromptBasedDRLexp1}
\end{figure}

The \promptTR attempts to fit the \conversion transfer function $h_\theta{(s^{\prime})}$ and transfer the representation from the target environment to the source one. 
It consists of four convolutional layers with kernel size of 4x4 and stride of 2 followed by a linear layer. 
We denote the learned optimal transfer function as ${h^{*}_\theta}{(s^{\prime})}$.
We use expert knowledge of imitation learning to initialize the weights of prompt-transformer in step~1, and optimize them by DRL in step~2 for better performance in the target environments.
 
\subsection{Step 1: \promptTRUp Initialization by Imitation Learning}
\label{oneStageIM}
We use imitation learning to initialize \promptTR by relying on expert data in the target environment. This allows \promptTR to fit the initial \conversion transformation function $h_\theta{(s^{\prime})}$. 
We use demonstration data denoted as $\left\{(s_{1}^{\prime}, a_{1}^{\prime}), (s_{2}^{\prime}, a_{2}^{\prime}), \ldots, (s_{n}^{\prime}, a_{n}^{\prime})\right\}$ given by the expert to train the policy $\pi_{\theta}\left(a^{\prime} \mid s^{\prime}\right)$, where $s^{\prime}$ is the state and $a^{\prime}$ is the behavior in the target environment.
Imitation learning discretizes the action space and uses a classifier (e.g., softmax output and cross-entropy loss). The training loss function is: 
\begin{equation}
\begin{aligned}
\label{eq_drl_pre}
    \mathcal{L}_{\text {imitation}}^{Step1}=-\sum_i a_i^{\prime} \cdot \ln \pi_{\theta}\left(\cdot \mid  h_\theta{(s_{i}^{\prime})}\right)
\end{aligned}
\end{equation}
Specifically, Fig.~\ref{mainArc} b) shows the pipeline of step~1, which includes the \promptTR, policy layers, action layers, and fully connected (FC) layers. The network structures of the policy and action layers are fixed and their weights, namely ${W_{policy}^{DRL}}$ and ${W_{action}^{DRL}}$, are from pre-trained models in source environments.
These weights are frozen during training.
When training in target environments in step~1, the network uses a small amount of expert data to initialize the \conversion transformation function $h_\theta{(s^{\prime})}$ with weights ${W_{prompt}^{Step1}}$. This function is used for step~2 to continue optimizing and try to reach the optimal value $h_\theta^{*}{(s^{\prime})}$.

\subsection{Step 2: \promptTRUp Optimization by Proximal Policy Optimization}
\label{twoStageDRL}

\begin{figure}[t]
    \centering
    \includegraphics[width = 0.98\linewidth]{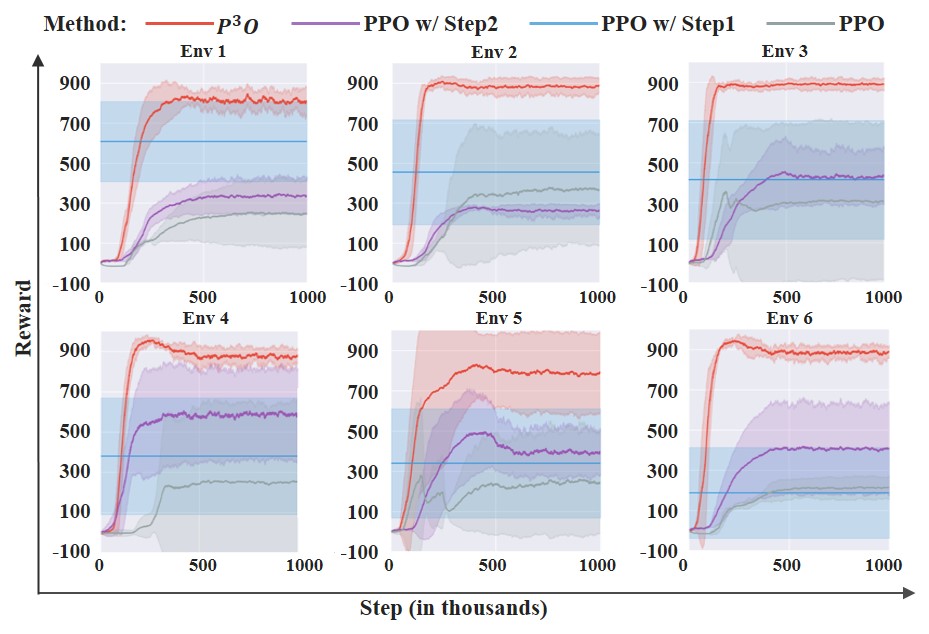}
    \caption{The comparison of the reward curves for the training by PPO only (in \textcolor[RGB]{140,155,159}{gray}), PPO with step~2 (in \textcolor[RGB]{138,67,168}{purple}), PPO with step~1 (in \textcolor[RGB]{40,129,210}{blue}), and \pbppo (in \textcolor[RGB]{224,51,46}{red}), {\em i.e.}, PPO with both step~1 and step~2.
   }\label{Ablationexp1}
    
\end{figure}

\begin{figure}[t]
	\centering
    \includegraphics[width = 0.97\linewidth]{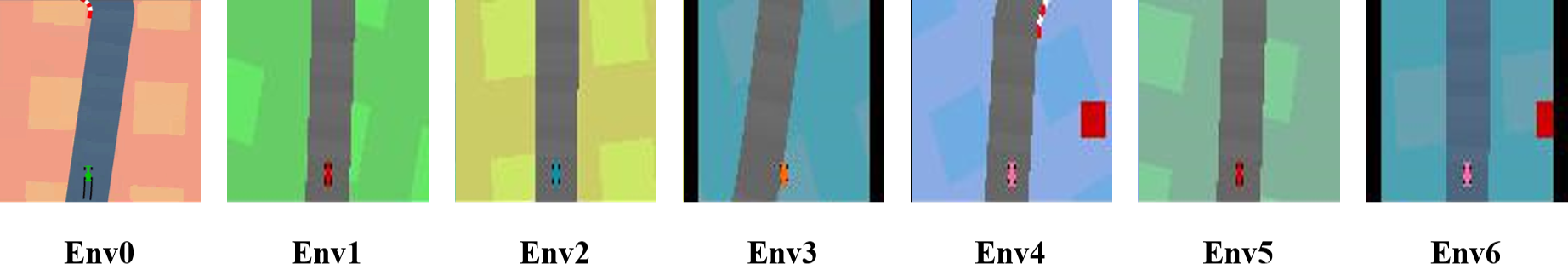}
	\caption{\label{fig:CarRacing_env}The source environment $Env0$ and its variant environments $Env1\sim Env6$ in the OpenAI CarRacing video game.}
\end{figure} 

In step~1, we obtain an efficient \promptTR initialization $h_\theta{(s^{\prime})}$. However, even when the \conversion of target environments is transformed by $h_\theta{(s^{\prime})}$ and input into the pre-trained model, the network still cannot obtain a score close to that in source environments. This indicates that $h_\theta{(s^{\prime})}$ is not yet capable of transforming the \conversion enough for the pre-trained model to recognize them. 
To improve the performance of the pre-trained model in the target environment, we use the PPO algorithm to further optimize the \conversion conversion function $h_\theta{(s^{\prime})}$. This function converts the representations of the target environment to the source environment, allowing the pre-trained model to achieve optimal performance.
The loss function of the policy network is:
\begin{equation}
\begin{split}
\label{st2_policy}
    \mathcal{L}_{\text {policy }}^{Step2}=&-\hat{\mathbb{E}}\left[\min \left(r_{\theta} \hat{A}{(h_\theta{(s^{\prime})},a^{\prime})},\right.\right.\\ 
    &\left.\left.\operatorname{clip}\left(r_{\theta}, 1-\epsilon, 1+\epsilon\right) \hat{A}{(h_\theta{(s^{\prime})},a^{\prime})}\right]\right.
\end{split}
\end{equation}
The loss function of the value network is:
\begin{equation}
\begin{aligned}
\label{st2_vaule}
    \mathcal{L}_{\text {value }}^{Step2}=\frac{1}{n} \sum_{i=1}^{n}(V(h_\theta{(s^{\prime})})-R)^{2}
\end{aligned}
\end{equation}
Fig.~\ref{mainArc} c) shows the pipeline of step~2, which includes the \promptTR, policy, action, and value layers. Their structures and weights are the same as those trained in the source environment.
The \promptTR (${W_{prompt}^{step1}}$) are pre-trained in step~1.
The policy, action and value layers are frozen during training.
The \conversion obtained from the transformation function in the target environment is $\boldsymbol{s}^{*}=h_\theta{(s^{\prime})}$. When the reward reaches its maximum value, the learned weights ${W_{prompt}^{step2}}$ of \promptTR are regarded as the optimal ones.

\subsection{DRL Predicting in the Target Environment}
\label{Predictinig}
In this stage, we use the optimal \promptTR to convert the \conversion of the target environment which is input into the pre-trained model to predict the action of the target environment. The prediction action of this stage can achieve similar or better scores in the target environment than the pre-trained model in the source environment, as shown in Fig.~\ref{mainArc} d).

\section{Experiments}
\setlength{\tabcolsep}{2.8pt}
\begin{table}[t]
\centering
\scriptsize
\caption{\label{tab:main}The comparison of transfer performance between \pbppo, PPO, PPO-FT, LUSR, and WAPPO.
The ratio in parentheses, which shows the transfer ability, is obtained by dividing the score for environments of the target scenario and the score for the source scenario.}
\begin{tabular}{cc|ccccc}
    \hline
    
    \hline
     \multicolumn{2}{c|}{\multirow{2}{*}{Envs}} & \multicolumn{5}{c}{Average Score (Transfer Ratio) $\uparrow$ }\\
    & & \textbf{\pbppo} (Ours) & PPO & PPO-FT  & LUSR & WAPPO \\
    \hline
    Sour. &Env0 & 816.2 & 432.5 & 816.2 & 825.5 & 684.9 \\ 
    \hline
    \multirow{5}{*}{Tar.}& Env1 & \textbf{887.4 (1.09)} & 176.8 (0.41) & 337.7 (0.41) & 808.3 (0.98) & 711.3 (1.04) \\ 
    &Env2 & \textbf{862.6 (1.07)} & 304.3 (0.70) & 84.3 (0.10) & 830.1 (1.01) & 659.8 (0.96) \\ 
    &Env3 & \textbf{854.4 (1.05)} & 288.8 (0.67) & 346.4 (0.42) & 791.3 (0.96) & 425.7 (0.66) \\ 
    &Env4 & \textbf{888.1 (1.09)} & 207.1 (0.48) & 327.7 (0.46) & 833.0 (1.01) & 615.1 (0.90)\\ 
    &Env5 & \textbf{891.4 (1.09)} & 177.8 (0.41) & 387.4 (0.47) & -58.5 (-0.07) & 579.7 (0.85) \\
    &Env6 & \textbf{880.8 (1.08)}& 128.9 (0.30)& 240.6 (0.30) & -0.2 (0.00) & -19.1 (-0.03)  \\ 
    \hline
    
    \hline
\end{tabular}
\end{table}

\subsection{\label{Implementation}Implementation Details}
\parsection{Environments}
Our experimental platform is the widely used OpenAI CarRacing video game~\cite{brockman2016openai}. Fig.~\ref{fig:CarRacing_env} illustrates the environments for seven different scenarios, where $Env0$ is the source environment, and $Env1 \sim Env6$ are the target environments.

\parsection{DRL Pre-training in the Source Environment}
We use PPO to learn a high-performance policy model in source environments of $Env0$ and consider it as the pre-trained model. The input of the PPO network is the observation data of a single frame, \textit{i.e.}, the game screen of the CarRacing environment, and the outputs are the actions, which are sampled from a Gaussian distribution. The learning rate in the pre-training stage is $1.0 \times {10^{ - 3}}$. We train the network with batch size = $128$, image size = ${96}\times {96}$. 

\parsection{Step~1: \promptTRUp Initialization by Imitation Learning}
In each target environment, we collect 4 sets of expert data, each containing 1500 pairs of data that complete a full lap of the track.
Next, we train the imitation learning network with a learning rate $1.0 \times {10^{ - 3}}$ using the small amount of expert data collected in each scenario.
After training, we save the parameters of \promptTR trained in the six environments of step~1 for step~2.

\parsection{Step~2: \promptTRUp Optimization by Proximal Policy Optimization}
The \promptTR in step~2 uses the weights obtained in step~1, and the weights of policy, action, and value layers are loaded from the pre-trained model. During step~2, the parameters of the policy, the action, and the value layers are also frozen, and only the \promptTR is learned. 
The learning rate used in this step is $2.0 \times {10^{ - 4}}$.

\subsection{Comparison of Transfer Performance}
We compare the transfer performance of \pbppo with four baselines: PPO~\cite{schulman2017proximal}, PPO with fine-tune (PPO-FT)~\cite{schulman2017proximal}, the prior state-of-the-art for the model-free policy gradient algorithm for DRL, LUSR~\cite{DBLP:conf/aaai/XingNCZNK21}, the prior state of the art for domain adaptation in DRL, and WAPPO~\cite{Roy_Konidaris_2021}, the prior state of the art for visual transfer in DRL. We report the reward for each task. In addition, we also compare PPO with Random Policy (Random).
Note that the LUSR and Random methods use the pre-trained model and the random initialization model of the source environment, respectively, so we report the target performance in Fig.~\ref{fig:PromptBasedDRLexp1} as horizontal lines. The light-colored area of the curve in Fig.~\ref{fig:PromptBasedDRLexp1} and Fig.~\ref{Ablationexp1} is the confidence interval.

\setlength{\tabcolsep}{9.5pt}
\begin{table}[t]
\centering
\scriptsize
\caption{\label{tab:step}The comparison of training efficiency between \pbppo, PPO, PPO-FT, LUSR and WAPPO. 
`F' denotes that the algorithm fails to converge in the scenario.}

\begin{tabular}{c|ccccc}
    \hline
    
    \hline
     \multicolumn{1}{c|}{\multirow{2}{*}{Envs}} & \multicolumn{5}{c}{Convergence Step $\downarrow$}\\
    & \pbppo (Ours)& PPO & PPO-FT  & LUSR & WAPPO \\ 
    \hline
    Env1 & \textbf{470k} & 880k & 810k & - & 2320k  \\ 
    Env2 & \textbf{310k} & 780k & 1290k & - & 2420k \\ 
    Env3 & \textbf{210k} & 660k & 1350k & - & 2290k  \\ 
    Env4 & \textbf{440k} & 570k & 1960k & - & 2560k \\ 
    Env5 & \textbf{620k} & 840k & 990k & - & 2110k \\
    Env6 & \textbf{410k}& 670k & 2000k & - & F  \\ 
    \hline
    
    \hline
\end{tabular}
\end{table}

\parsection{Algorithm Score}
Fig.~\ref{fig:PromptBasedDRLexp1} shows the comparison results of the reward for the training by \pbppo, PPO, LUSR, WAPPO, and Random methods in $Env1\sim Env6$.
The comparison results show that the reward of \pbppo after convergence in the graph exceeds that of other methods in all environments. 
Table ~\ref{tab:main} summarizes the average score and transfer ratio of PPO, PPO-FT, LUSR, WAPPO and \pbppo in the target environment.
\pbppo outperforms other algorithms in terms of average score and transfer ratio. The average transfer ratio of \pbppo is 1.66 times and 1.48 times that of LUSR and WAPPO respectively.
\pbppo's transfer ratios in all target environments are over 1.0, indicating that \pbppo can leverage the knowledge from the pre-trained model to achieve similar or even higher scores compared to the source environment. This also shows that \promptTR plays a consistent role in visual \conversion transfer across different environments, while other algorithms may lose their effect in some environments.

\parsection{Algorithm Efficiency}
Table ~\ref{tab:step} shows the convergence steps of several algorithms. LUSR only uses the pre-trained model, so it is labeled as `-'. Among them, \pbppo converges in fewer steps and obtains the highest score than other algorithms.  Specifically, PPO, PPO-FT, and WAPPO took 1.79, 3.42, and 5.71 times as many steps as \pbppo to complete convergence, with lower scores. This shows that \pbppo is more efficient and can complete the visual \conversion transfer task faster. Our \pbppo (red curve) in Fig.~\ref{fig:PromptBasedDRLexp1} reaches the highest point faster than other methods in all experiments, further proving its efficiency.
\pbppo has higher algorithm efficiency, which enables it to complete the migration of policy at a lower cost and in a shorter time when encountering a new environment.

\parsection{Algorithm Confidence Interval}
\pbppo has a tighter confidence interval than other algorithms, as shown in Fig.~\ref{fig:PromptBasedDRLexp1}, indicating that its results are more consistent and stable. In our experiments, \pbppo consistently outperforms other algorithms when the environment changed unexpectedly, showing its superior stability and adaptability. 

\subsection{Ablation Studies}
In order to study the influence of each part of the \pbppo algorithm on the experiment, we set up four groups of experiments: PPO, PPO w/ step~1, PPO w/ step~2, PPO w/ both steps, \emph{i.e.}, \pbppo. The experimental results are shown in Fig.~\ref{Ablationexp1} and Table~\ref{tab:ablationScore}.

\setlength{\tabcolsep}{4pt}
\begin{table}[t]
\scriptsize
\begin{center}
\caption{\label{tab:ablationScore}
Ablation analysis on \pbppo in $Env1\sim Env6$.}

\begin{tabular}{ccc|cccccc}
    \hline
    
    \hline
    \multirow{2}{*}{PPO} & \multirow{2}{*}{Step\ 1} & \multirow{2}{*}{Step\ 2} & \multicolumn{6}{c}{\text{Average\ Score $\uparrow$}}\\
    & & & Env1 & Env2 & Env3& Env4 & Env5 & Env6\\ 
    \hline
    \checkmark & & & 176.8 & 304.3 & 288.8 & 207.1 & 177.8 & 128.9 \\ 
    \checkmark & & \checkmark& 246.5 \textcolor[RGB]{224,51,46}{↑} & 173.2\textcolor[RGB]{85,199,84}{↓} & 393.8\textcolor[RGB]{224,51,46}{↑} & 570.0\textcolor[RGB]{224,51,46}{↑} & 466.5\textcolor[RGB]{224,51,46}{↑} & 329.3\textcolor[RGB]{224,51,46}{↑} \\ 
    \checkmark &\checkmark & & 607.5\textcolor[RGB]{224,51,46}{↑} & 453.9\textcolor[RGB]{224,51,46}{↑} & 414.1\textcolor[RGB]{224,51,46}{↑} & 379.5\textcolor[RGB]{224,51,46}{↑} & 338.9\textcolor[RGB]{224,51,46}{↑}& 186.7\textcolor[RGB]{224,51,46}{↑} \\ 
    \checkmark &\checkmark &\checkmark & 887.4\textcolor[RGB]{224,51,46}{↑} & 862.6\textcolor[RGB]{224,51,46}{↑}& 854.4\textcolor[RGB]{224,51,46}{↑}& 888.1\textcolor[RGB]{224,51,46}{↑} & 891.4\textcolor[RGB]{224,51,46}{↑} & 880.8\textcolor[RGB]{224,51,46}{↑}\\ 
    \hline
    
    \hline
\end{tabular}
\end{center}
\end{table}
\setlength{\tabcolsep}{1.4pt}

\parsection{The Impact of Step~1}
Step~1 uses imitation learning to guide the learning of \promptTR effectively.
When comparing PPO and PPO with step~1, after introducing step~1, the rewards and average scores in Fig.~\ref{Ablationexp1} and Table~\ref{tab:ablationScore} have a small range of improvement. It shows that step~1 can provide guidance for the learning of \promptTR, and \promptTR has the ability to transfer visual representation.
When comparing PPO with step~2 and PPO with step~1 and step~2 (\pbppo), the lack of step~1 guidance will lead to a large decrease in rewards and average scores in Fig.~\ref{Ablationexp1} and Table~\ref{tab:ablationScore}, which also shows that step~1 provides effective initialization for \promptTR is an essential part of the experiment.

\parsection{The Impact of Step~2}
Step~2 is responsible for the continuous optimization of \promptTR.
After initializing the \promptTR in step~1, we show the rewards and average scores in Fig.~\ref{Ablationexp1} and Table~\ref{tab:ablationScore} to verify the effect of step~2. 
The comparison of experiments between PPO with step~1 only and PPO with both steps (\emph{i.e.}, \pbppo) shows that the reward and the average score have significantly improved after introducing step~2.

\section{Conclusions}
In this work, we propose \pbppo, a three-stage DRL algorithm that transfers visual representations to quickly adapt to target environments by directly reusing the knowledge of pre-trained models from the source environment.
Inspired by the prompting method in NLP, we design a representation transfer module \promptTR and use the three-stage DRL algorithm to optimize \promptTR so as to realize the representation transfer task from the target environment to the source environment.
Experiments on the OpenAI CarRacing video game show that \pbppo reaches the state-of-the-art in visual \conversion transfer. 
In future work, we will extend our algorithm to the cross-environment and cross-modal tasks for the high-performance DRL method driven by prompting.   
\bibliographystyle{IEEEtran}
\bibliography{icme_ieee_tran_ref}

\end{document}